# Development and evaluation of automated localisation and reconstruction of all fruits on tomato plants in a greenhouse based on multi-view perception and 3D multi-object tracking

David Rapado-Rincon, Eldert J. van Henten, Gert Kootstra

**Abstract**

The ability to accurately represent and localise relevant objects is essential for robots to carry out tasks effectively. Traditional approaches, where robots simply capture an image, process that image to take an action, and then forget the information, have proven to struggle in the presence of occlusions. Methods using multi-view perception, which have the potential to address some of these problems, require a world model that guides the collection, integration and extraction of information from multiple viewpoints. Furthermore, constructing a generic representation that can be applied in various environments and tasks is a difficult challenge. In this paper, a novel approach for building generic representations in occluded agro-food environments using multi-view perception and 3D multi-object tracking is introduced. The method is based on a detection algorithm that generates partial point clouds for each detected object, followed by a 3D multi-object tracking algorithm that updates the representation over time. The accuracy of the representation was evaluated in a real-world environment, where successful representation and localisation of tomatoes in tomato plants were achieved, despite high levels of occlusion, with the total count of tomatoes estimated with a maximum error of 5.08% and the tomatoes tracked with an accuracy up to 71.47%. Novel tracking metrics were introduced, demonstrating that valuable insight into the errors in localising and representing the fruits can be provided by their use. This approach presents a novel solution for building representations in occluded agro-food environments, demonstrating potential to enable robots to perform tasks effectively in these challenging environments.

**List of Nomenclature**

| | |
|---|---|
| DoF | Degrees of Freedom |
| FN | False Negative |
| FNA | False Negative Association |
| FP | False Positive |
| FPA | False Positive Association |
| HOTA | Higher Order Tracking Accuracy |
| IoU | Intersection Over Union |
| LiDAR | Light Detection and Ranging |
| MAPE | Mean Absolute Percentage Error |
| MOT | Multi-Object Tracking |
| MPE | Mean Percentage Error |
| NBV | Next-Best View |
| NSF | Non-valid Sphere Fitting |
| RD | Rejected Detection |
| RGB | Red Green Blue |

.

# 1. Introduction

An increasing food demand combined with the existing labour shortages have become one of the most important challenges in the agro-food industry (Ince Yenilmez 2015). Robotic systems are recognised as one of the main technologies that will play a key role in solving these issues (Bogue 2016). Consequently, agro-food robotics has continued to grow as a field for more than three decades. However, most agro-food robots are still far from being an economically feasible alternative due to the challenging nature of agro-food environments (Roldan et al. 2017; Kootstra et al. 2020). These challenges are the presence of heavy occlusions (Roldan et al. 2017); and the wide variation in environment conditions, cultivation systems, and object and plant positions, sizes, shapes and colours (Bac et al. 2014; Zhao et al. 2016). These challenges form a limiting factor in the performance of most agro-food robotic systems (Montoya-Cavero et al. 2022).

Accurate representation and localisation of relevant objects are crucial for robots to perform their tasks effectively. Developing a comprehensive representation is not only essential for addressing these challenges but also for enhancing the overall performance of agro-food robots. This representation, known as a world model (Crowley 1985; Elfring et al. 2013), enables robots to better understand and navigate their environment, particularly when dealing with occlusions. Traditional approaches, where harvesting robots simply capture an image, attempt to perform an action, and then discard the previously retrieved information, struggle in the presence of occlusions. As a result, these traditional approaches experience a significant impact on their performance in highly occluded agro-food environments, where localising relevant objects can be a considerable challenge (Arad et al. 2020).

Methods using multi-view perception have the potential to effectively deal with occlusions. However, to effectively integrate information from multiple viewpoints, they require a world model that can integrate and extract information from the viewpoints, enabling robots to better handle occlusions and adapt to changing environments. Another motivation for world models lies in the ability to become more flexible to deal with different objects of interest to allow agro-food robots to deal with different tasks and environments. The objects of interest can vary depending on the task and environment. For instance, different parts of a plant may be of importance to the robot when monitoring the plant status versus harvesting ripe fruits.

Additionally, different environments can have different objects of interest, such as various fruits for different crops. The agro-food industry requires systems that can adapt to these varying conditions easily, which can be achieved through the development and application of versatile world models, which form the bridge between raw sensor data and a robust abstract representation of the relevant aspects of the task environment.

Most current agro-food robotics research aims to improve the perception capabilities of the systems, which increases the robots′ understanding of a scene. In recent years, most of the research has been done in increasing detection and localisation performance, especially through deep learning (Blok et al. 2022; Ruigrok et al. 2020). Although these methods have shown to be able to deal with variation to some extent, the challenge of occlusion is not adequately addressed (Montoya-Cavero et al. 2022; Kootstra et al. 2021). Barth et al. (2016) and Lehnert et al. (2019) showed that multi-view perception can be successfully used to reduce occlusions. Barth et al. (2016) developed an eye-on-hand architecture which was able to create a 3D geometrical reconstruction in the form of a point cloud using visual simultaneous localisation and mapping (SLAM). However, SLAM algorithms are limited in the description of the objects that they can generate since they only use geometrical features. Therefore, this limits the possible interactions of the robot with them (Nicholson et al. 2019). Lehnert et al. (2019) developed a visual servoing to find the next-best-view and remove occlusions in agro-food environments. They showed how their method, when implemented in a 6-degrees-of-freedom (DoF), can potentially reduce occlusions generated by plants. However, in their experiments they used only a single object, which considerably simplifies the representation needs.

Several authors have shown how a world model based on objects with attributes that are updated over time can be used to deal with the challenges of complex and occluded environments (Elfring et al. 2013; Persson et al. 2020; Wong et al. 2015). These approaches have in common the use of multi-object tracking (MOT) to link upcoming noisy and uncertain measurements from the sensors of the robot with the representation of the objects that generated those measurements. Unlike SLAM-based representations, MOT-based representations are not only geometrical as they are based on objects which can have different semantic properties and attributes. Furthermore, MOT has the potential to be useful when dealing with occlusions since they allow for merging of sensor information from different viewpoints while the robot moves, even in complex scenes where many objects are present. This is important as new viewpoints can give new information about areas that appeared occluded to the robot before. This means that they can couple well with active perception algorithms, such as next-best view (NBV) methods (Lehnert et al. 2019; Wu et al. 2019; Burusa et al. 2022), to offer efficient mechanisms to explore and represent highly occluded agro-food environments.

MOT methods have been recently used in agro-food robotic systems (Halstead et al. 2018, 2021; Kirk et al. 2021; Smitt et al. 2021). Nevertheless, their objectives were to perform specific tasks like plant phenotyping and monitoring, and were not aiming for a generic representation to improve the capabilities of an autonomous robotic system in tasks like harvesting and crop maintenance. Halstead et al. (2018) developed a deep learning-based detection system to detect and assess the ripeness of peppers, and used it to feed an image-based Intersection over Union (IoU) tracking system. They used the IoU between the bounding boxes of tracks and detections to generate a cost matrix based on which the data association was performed using the Hungarian algorithm. Their goal was fruit counting and ripeness estimation. The work of Halstead et al. (2018) was extended by Smitt et al. (2021) to autonomously count fruits over a greenhouse plant row. They made use of the 3D data obtained from the camera mounted in the robotic system to re-project the bounding boxes of tracked objects onto the following frames, then the tracking was performed by the previously mentioned IoU-based system. Halstead et al. (2021) further extended the work of Smitt et al. (2021) to different cropping systems, like arable farmland, and evaluated the performance on sugar beet plant counting. Kirk et al. (2021) also developed an IoU-based tracking system assisted by deep learning, inspired by the work done by Wojke et al. (2017). They used their method to count strawberries in different ripeness stages.

The above-mentioned studies of Halstead et al. (2018); Kirk et al. (2021); Smitt et al. (2021) and Halstead et al. (2021) have the use of an autonomous systems with only one DoF that moves and collects data in a line parallel to the plant row or soil in common. This could be enough for monitoring applications in cropping systems where the objects of interest grow relatively wide spaced, as is the case for strawberries or already pruned bell peppers. However, it might not be optimal in cropping systems where fruits or plants grow very close to each other, like tomatoes in a truss. Furthermore, the 1DoF motion provides only limited capabilities to deal with occlusions as their feasible motion can be far from the next-best-view. In addition, all aforementioned research used tracking algorithms to perform counting of fruit parts, yet they evaluated the performance over a total count error without studying the actual tracking performance of the algorithms. This is considered a key aspect of the evaluation of a representation algorithm based on object tracking, as even though the total count of objects might seem correct, there could be a multitude of unseen errors, such as switching tracks between objects or false positives and negatives that cancel each other out. These cannot be assessed using only the count error. This means that good counting of objects does not directly correspond with perfect tracking, and therefore, with a good representation of the objects in the environment. If left without evaluation, the representation algorithm might perform worse on new data or in new environments. In addition, these errors might not be an issue for tasks like yield prediction, however, they will be an important issue for more advanced tasks that require a more accurate representation, such as harvesting or monitoring of crop development.

In this paper, a generic method for building a world model for robots, which can be used in different tasks and in highly occluded environments, is presented. The improvement of the representation and localisation capabilities of agro-food robotic systems could help address the critical challenges facing the agro-food industry and drive progress towards the development of effective and economically feasible robotic solutions. The contributions are as follows:

- A method that can be used by robots to build accurate 3D object-centric representations of the environment. The representation is maintained and updated from multiple viewpoints over several time steps using a 3D MOT algorithm based on a Kalman Filter and the Hungarian algorithm.
- A real-world test and evaluation of the proposed method in a tomato greenhouse using a 6DoF eye-in-hand robotic arm.
- A novel analysis of the performance of the method using state-of-the-art MOT metrics, which allows a better understanding of the different types of errors that affect MOT algorithms in agro-food environments.

The data collection and labelling process, tracking algorithm, and evaluation metrics are presented in Section 2. The results of the experiments are shown in Section 3 and discussed in Section 4. Section 5 contains the conclusions drawn from the research and recommendations for future work.

## 2. Materials and Methods

### 2.1. Robotic set-up

An ABB IRB1200 robotic arm with an Intel Realsense L515 LiDAR camera attached to the end-effector was utilised. The camera was mounted to the end-effector of the robot. Eye-on-hand calibration was performed to get a correct estimation of the transformation between the camera frame and the robot frame. The whole system was controlled through the robot operative system (ROS) (Quigley et al. 2009). The robot arm was then mounted on a cart designed to move on the rails in between rows present in a Dutch tomato greenhouse to allow collection of data of multiple plants. Figure 1 shows an example of the data collection scene in the greenhouse.

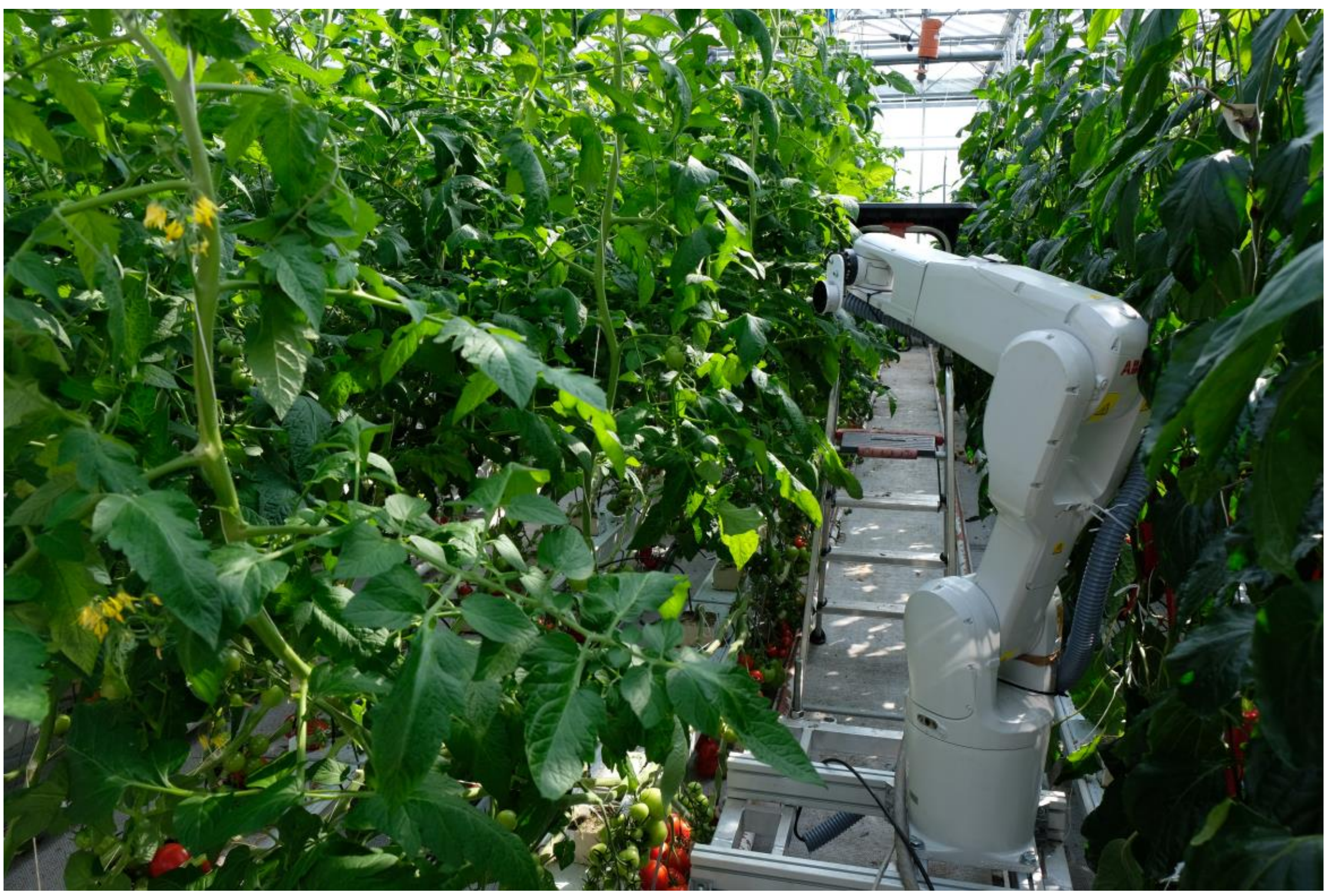

**Figure 1.** The robot standing in front of a tomato plant in a greenhouse. The robot was attached to a platform that was able to move over the rails between plant rows. An Intel Realsense L515 camera was attached to the end effector of the ABB IRB1200 robot arm.

### 2.2. Data collection and annotation

Two datasets were used for two different purposes. First, a RGB image dataset was used to train and evaluate a deep learning detection algorithm. The tomatoes in the colour images were labelled with a pixel-level mask and a bounding box. This dataset consisted of 1180 images and was based on the 123 images collected by (Afonso et al. 2019), and 1057 new images obtained for this research from greenhouses at Wageningen University & Research from the varieties Merlice and Campari. Therefore, it contained images with tomatoes from different varieties, greenhouses and light conditions.

A second dataset was collected using the robot in order to evaluate the world model algorithm. The data was collected in a tomato greenhouse from Wageningen University & Research. The dataset consisted of 700 images from 100 viewpoints of seven tomato plants recorded following a semi-cylindrical path as shown in Figure 2. The path was divided into ten different height levels, and each level corresponded to a semicircle path with ten different viewpoints distributed evenly over the path. The stem was considered the centre of the semi-cylinder and was assumed to be approximately at 60 cm in front of the robot origin of coordinates. The radius of the semi-cylinder was 30 cm. Starting from the first viewpoint, the robot stops at each viewpoint to collect the required data, and then moves to the next viewpoint. Seven different plants, from the variety Santiana (not present in the detection training dataset), were recorded. For each viewpoint, a 960x540 resolution colour image and associated structured point cloud were recorded, as well as the camera pose relative to the robot base. Figure 3 shows an example of two viewpoints of the same plant. The tomatoes on every frame's colour image were labelled using a bounding box and tracking ID as shown in Figure 4. This type of labelling allowed us to know which tomatoes are present and how many tomatoes in total have been seen in each frame.

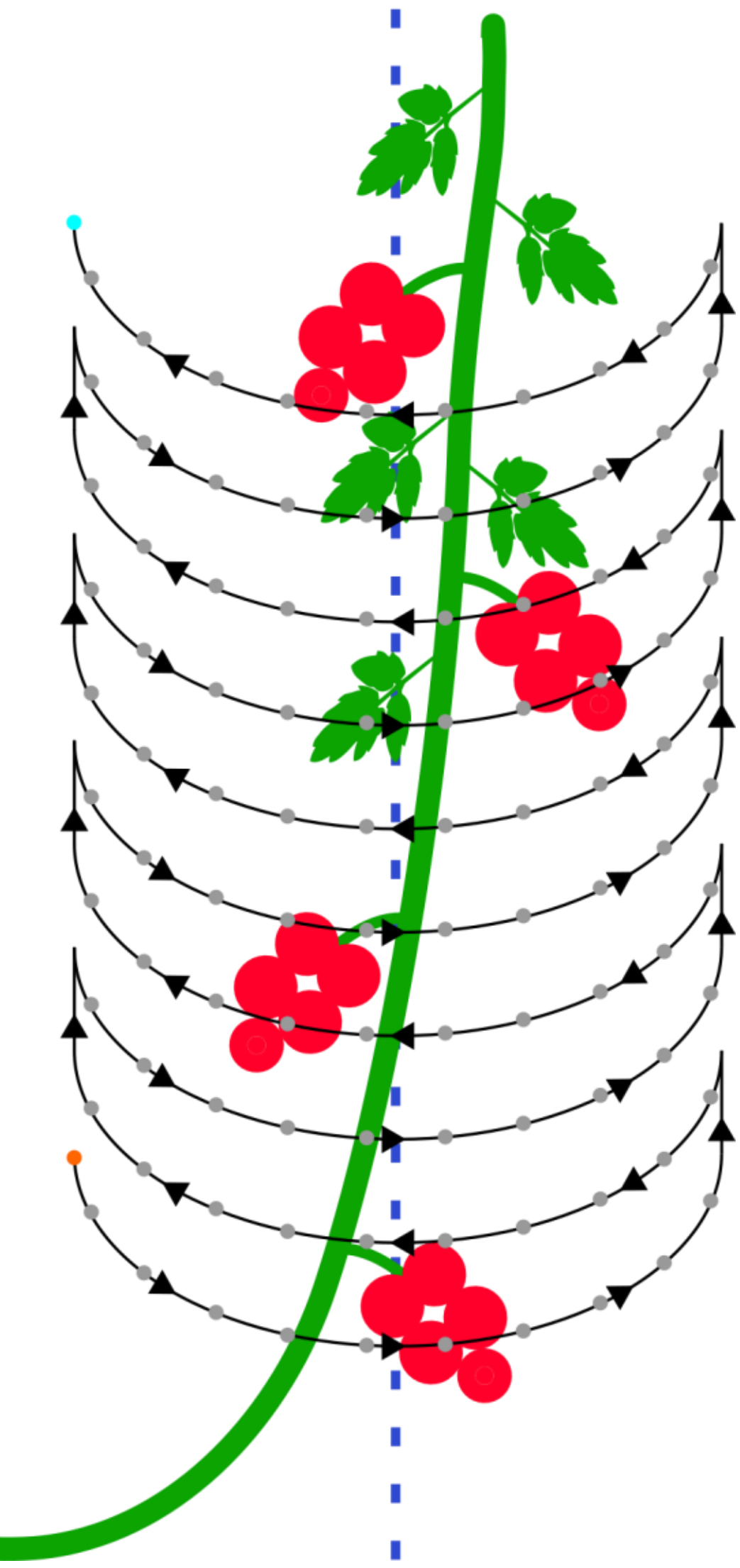

**Figure 2**. Example of the semi-cylinder path used to collect the dataset used to evaluate the accuracy of the representation. It was built using ten different heights, each with ten evenly distributed viewpoints, represented by the grey dots. At every viewpoint, the camera was pointing towards the centre of the semicircle of that height level (blue dashed line).

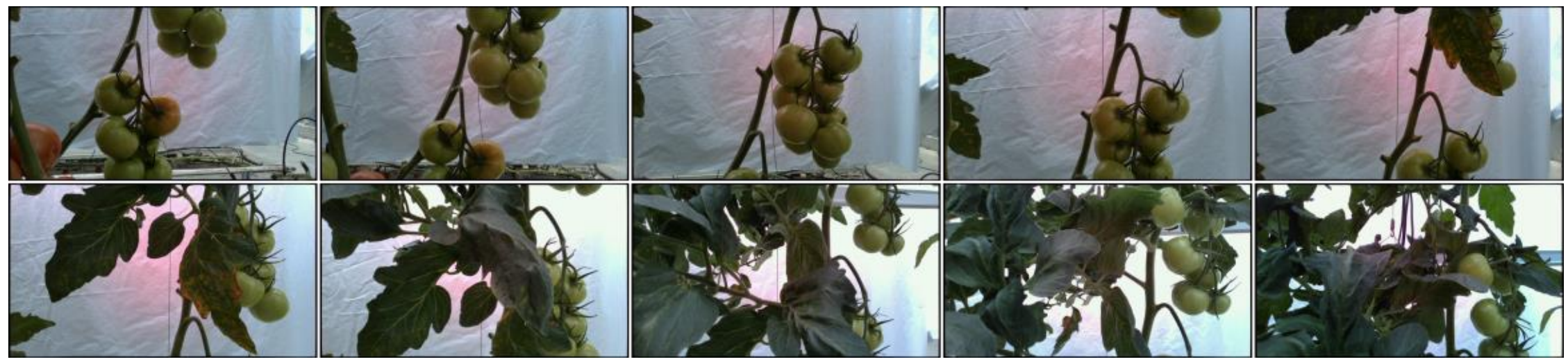

**Figure 3**. Examples of viewpoints of a plant at each one of the tenth heights. The upper part of the plant illustrates a more complex scene due to occlusions by leaves.

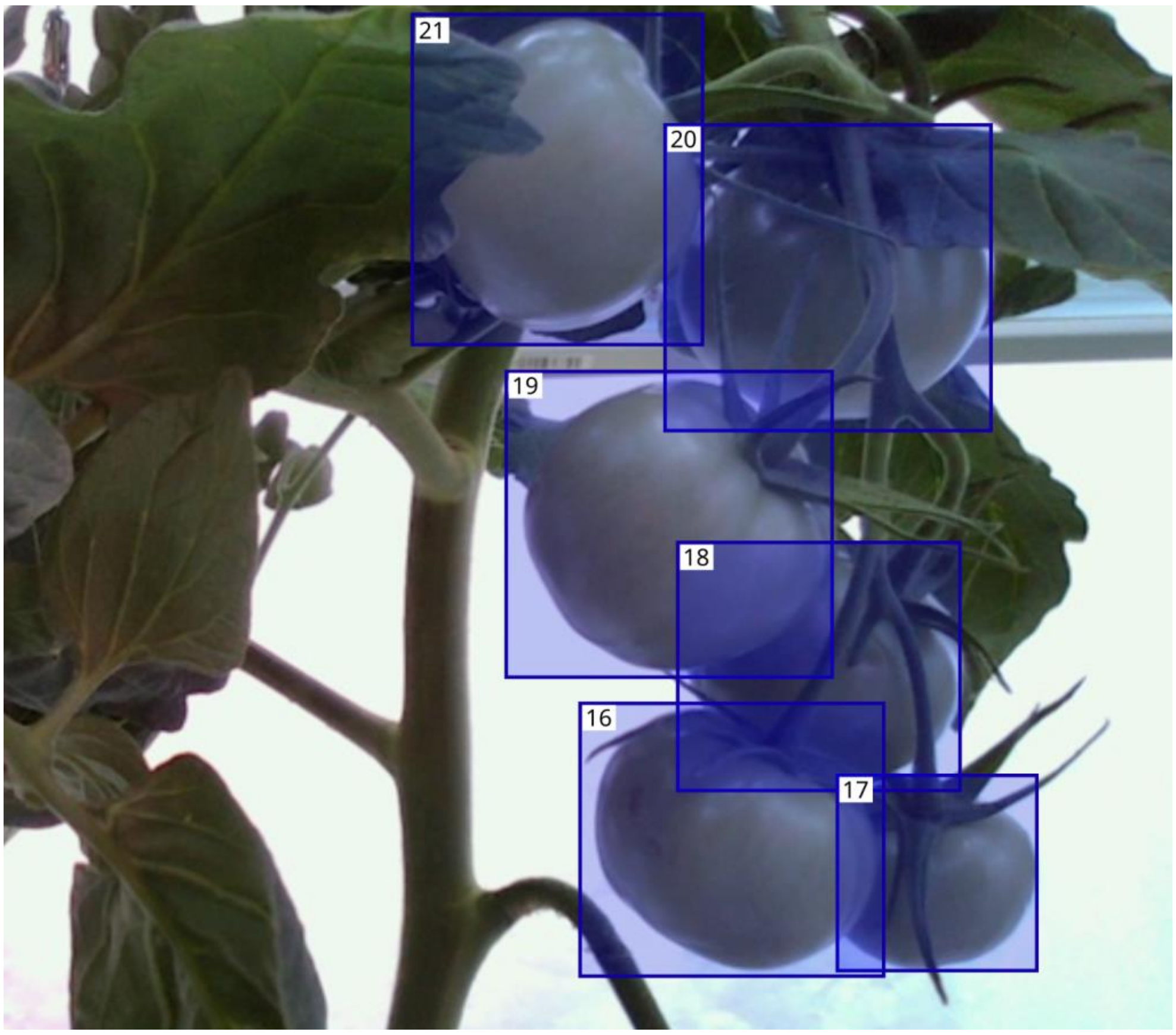

**Figure 4**. Example of a frame's colour image where the tomatoes are annotated using bounding boxes and tracking IDs.

### 2.3. Data pre-processing and object detection

At each time step, the detection algorithm depicted in Algorithm 1 detected objects, generating a set of $M$ object detections at time $t$. These detections were stored in $D_t = d_t^1, d_t^2, \dots, d_t^M$, where each $d_t^j$ included the 3D position of the detection with respect to the robot coordinate system $p_{rob,t}^j$, class $c_t^j$, bounding box $b_t^j$, and mask $m_t^j$. The position was defined as a Multivariate Gaussian distribution, $p^j_{rob,t} = \mu^j_{rob,t}, \Sigma^j_{rob,t}$, where the mean $\mu^j_{rob,t} \in \mathbb{R}^3$ and covariance $\Sigma^j_{rob,t} \in \mathbb{R}^{3x3}$. In the experiments, only tomatoes were used, so $c^j_t$ always corresponded to tomato.

---

**Algorithm 1** Detection Algorithm

---

**function** DETECTION(PointCloud, ColourImage, $T^{rob}_{cam}$)
  **for** $d^\alpha_t \in D^\alpha_t$ **do**
    TomatoPoints ← filterPointCloud($m^j_t$, PointCloud)
    **if** len(TomatoPoints) > 0 **then**
      $\mu^j_{cam,t}$, $r^j_t$ ← fitSphere(TomatoPoints)
      **if** $\mu^j_{cam,t}$ is not valid **AND** $r^j_t$ is not valid **then**
        $\mu^j_{cam,t}$ ← mean(TomatoPoints)
      $\mu^j_{rob,t}$, $\Sigma^j_{rob,t}$ ← transform($\mu^j_{rob,t}$, $\Sigma^j_{rob,t}$, $T^{rob}_{cam}$)
  **return** $D_t$

---

To obtain the object detections, the colour image was passed through a trained Mask R-CNN (He et al. 2017) algorithm. This generated a set of $M$ 2D detections at time $t$, denoted by $D^\alpha_t = d^{\alpha,1}_t, d^{\alpha,2}_t, \ldots, d^{\alpha,M}_t$. Here, $d^{\alpha,j}_t = c^j_t, b^j_t, m^j_t$ corresponded to the 2D detection $j$ at time $t$. Mask RCNN filters detections based on a defined confidence threshold. To assess the impact of this parameter, two commonly used confidence thresholds: 0.5 and 0.7, were utilised.

The transformation of 2D detections into 3D detections involved the filtering of the 3D points of each detected object from the whole point cloud using the mask $m^j_t$. However, the retrieved point clouds were often noisy and at times contained zero valid points, resulting in these cases being removed from the detection list. The remaining partial point cloud was passed to a sphere fitting algorithm to estimate the centre of every tomato $\mu^j_{cam,t}$ with respect to the camera coordinate system. Due to noise in the point cloud and small numbers of points, tomatoes whose centre $\mu^j_{cam,t}$ or radius $r^j_t$ fell outside of reasonable limits were filtered out. The radius was also defined as $\rho_{min} \leq r^j_t \leq \rho_{max}$ where $\rho_{min} = 1cm$ and $\rho_{max} = 5cm$. If a radius fell outside of these limits, it was considered the result of faulty sphere fitting, but its centre was still estimated as the average of all the 3D points assigned to it. The known camera pose with respect to the robot base was then used to calculate the transformation matrix $T^{rob}_{cam}$, which was used to transform the tomato position $p^j cam, t$ to the robot base coordinate frame $p^j_{rob,t}$.

Mask R-CNN detected tomatoes in the whole image, including tomatoes from other rows or plants. To ensure only the tomatoes of interest were selected, they were filtered out based on their transformed position $p_{rob,t}^{j}$ using thresholds measured at data recording time. Based on the spacing between the plants in the row, each axis $\mu_{rob,t}^{j}$ was restricted as

$$\begin{aligned} -0.2 \leq \mu_{rob,t}^{j}(x) \leq 0.2 \\ -0.8 \leq \mu_{rob,t}^{j}(y) \\ 0.4 \leq \mu_{rob,t}^{j}(z) \end{aligned} \qquad (1)$$

where $\mu_{rob,t}^{j}(x)$ corresponds to the axis parallel to the plants row, $\mu_{rob,t}^{j}(y)$ is the axis perpendicular to the plants row, and $\mu_{rob,t}^{j}(z)$ is the axis which represents the height with respect to the ground. Values are presented in meters. Since further steps are only performed in the robot coordinate system, $p_{rob,t}^{j}$ is abbreviated as $p_{t}^{j}$.

## 2.4. World model

The representation, or world model, of the robot's environment consisted of a set of $N$ object representations or tracks $O_t = \{o_t^1, o_t^2, \ldots, o_t^N\}$ where $o_t^i = \left(p_t^i, c_t^i, b_t^i\right)$ corresponded to a specific track with position $p_t^i$, class label $c_t^i$, and bounding box $b_t^i$ at time $t$. From a tracking perspective, objects in the world model are also referred as tracks. The position was represented as a Multivariate Gaussian distribution, $p_t^i = \{\mu_t^i, \Sigma_t^i\}$, where the mean, $\mu_t^i \in \mathbb{R}^3$, represented the most likely position of an object; and the covariance, $\Sigma_t^i \in \mathbb{R}^{3x3}$, the uncertainty on the position. The object position was represented in the world model coordinate system, which corresponds in the experiments to the robot coordinate system. The bounding box $b_t^i$ of an object was represented as the 2D bounding box of the last detection $d_t^j$ associated with that object. This was needed for the tracking evaluation metrics explained in Section 2.5.

Figure 5 shows how the world model was maintained and updated over time. At every frame, detections arrived from the object detection algorithm. They were passed on to the data association algorithm, which associated existing tracks with the detections. Then these associations were used to update the world model by updating the objects (or tracks). A prediction was made to project the tracks into the next time step (frame), which was then used in the next cycle by the data association algorithm together with detections from the next time step.

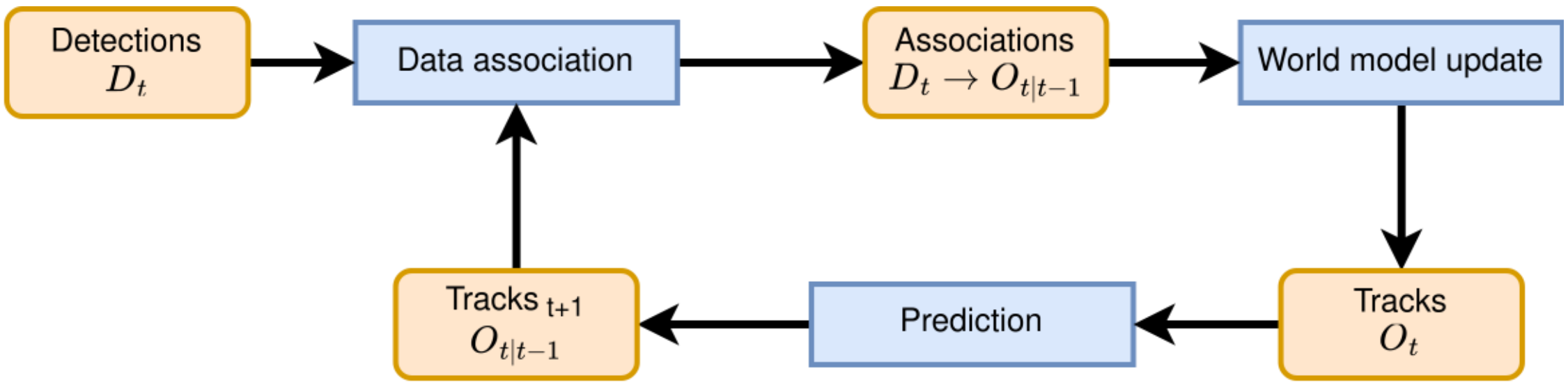

**Figure 5.** Process used to maintain and update the world model representation over time.

### 2.4.1. Data association

For every frame detected objects were assigned to the existing tracks using their predicted state from the previous time step as $D_t \to O_{t|t-1}$. This was done using the Hungarian algorithm. The method used a cost matrix, $C(i,j)$, where the cell i, j contained the cost of assigning the detection j to the track i, and outputted the combinations of i, j which minimised the total cost. In order to consider uncertainty, these cost were calculated as the square of the Mahalanobis distance

$$C(i,j) = \left(\mu_t^j - \mu_t^i\right) {\Sigma_t^i}^{-1} \left(\mu_t^j - \mu_t^i\right)^T \qquad (2)$$

where $\mu_t^i$ and $\Sigma_t^i$ defined the Gaussian distribution that represents the position of the track $o^i$, and $\mu_t^j$ corresponded to the mean position of the detection $d_t^j$. Unlikely associations were discarded by setting a threshold to this distance metric. Since the squared Mahalanobis distance follows a chi-square ($\chi^2$) distribution, this threshold was set to 7.82, a value which corresponds to the 0.95 quantile of the chi-square distribution with 3 degrees of freedom (Wojke et al. 2017). Associations whose $c(i,j)$ was larger than this threshold were discarded.

### 2.4.2. World model update

Once a track $o_t^i$ was associated with a detection $d_t^{i \to j}$, the attributes of the detection were used to update the object attributes. Object class $c_t^i$ and bounding box $b_t^i$ were updated with the detection class $c_t^j$ and bounding box $b_t^j$ respectively. The updated object position $p_t^i$ was calculated in a Kalman filter update step using the detection position $p_t^j$ and the predicted position of the object from the previous time step $p_{t|t-1}^i$.

New objects were initialised in the world model when a detection could not be associated with any existing object. In order to prevent initialising objects from false positive detections, new objects in the world model could be considered tentative during a certain number of frames, $n_{init}$. While confirmed tracks are never removed from the world models, tentative tracks are removed from it if they are not associated with any detection. During the experiments, the effect of different values for $n_{init}$ was assessed.

### 2.4.3. Prediction

In the prediction step, objects in the world model were propagated between one time step and the next. In this step, object class $c_t^i$ and bounding box $b_t^i$ were considered constant, and the object position $p_{t+1|t}^i$ was predicted using a Kalman Filter prediction step.

## 2.5. Experiments

To get an analysis of the performance of the world model representation algorithm two parts were evaluated:

1. **Pre-processing.** As shown in Algorithm 1, everything started with the data retrieved from the sensor. Due to the challenging light conditions of the environment, the point cloud was sometimes noisy and incomplete. This resulted in some tomato detections being rejected, or in larger inaccuracies on their estimated position when a sphere could not be fitted properly. To study the camera performance on the dataset, the percentage of detections which were rejected due to a lack of points and the percentage of detections whose predicted sphere fell outside of the established limits were calculated.

   The first step in the object detection algorithm was 2D object detection and instance segmentation through Mask R-CNN (He et al. 2017). Therefore, its performance played an important role in the performance of the system. The precision $P = \frac{T_p}{T_p+F_p}$ and the recall $R = \frac{T_p}{T_p+F_n}$ were evaluated, where true positives $T_p$, false positives $F_p$, and false negatives $F_n$ were defined by the IoU between the predicted and ground truth bounding boxes, using the standard IoU threshold of 0.5. This analysis was done over the two different used confidence thresholds, 0.5 and 0.7.

2. **World model representation**. To evaluate the performance of the world model algorithm, different values for the initialisation threshold $n_{init}$ were explored during the experiments. Besides, two different analyses were performed: counting and data association. The counting performance of the world model was evaluated similarly to the evaluation made in recent papers that used data association for monitoring and yield prediction on crops (Kirk et al. 2021; Smitt et al. 2021). The evaluation of the actual data association performance was performed using a state-of-the-art multi-object tracking metric, Higher Order Tracking Accuracy (HOTA) (Luiten et al. 2021).

   The metrics used to evaluate the count performance were the mean percentage error:

$$\text{MPE} = \frac{100}{n} \sum_{i=1}^{N} \frac{A_i - P_i}{A_i} \qquad (3)$$

and the mean absolute percentage error

$$\text{MAPE} = \frac{100}{n}\sum_{i=1}^{N}\frac{|A_i - P_i|}{A_i} \quad (4)$$

where $A_i$ is the actual value and $P_i$ is the predicted value, and $n$ is the number of frames. MPE gives an estimation of the counting accuracy which reflects under or over counting, while MAPE gives the relative absolute error on the counting.

Evaluating the accuracy on object count over a series of sequences, on its own, is not enough to assess the performance of a data association algorithm. A good counting performance can still hide plenty of association errors like ID switching, or false positives that counter false negatives. Therefore, the tracking performance of the algorithm was evaluated using HOTA, a standard multi-object tracking metric (Luiten et al. 2021). It is a metric that can be easily decomposed into the three main sources of errors for tracking algorithms: localisation, detection and association. Everything starts with the IoU threshold, $\alpha$, that is used to define true positives (TP), false positives (FP), and false negatives (FN) similarly to how a detection algorithm is evaluated. From it, the localisation accuracy can be measured as:

$$\text{LocA}_{\alpha} = \frac{1}{|\text{TP}_{\alpha}|}\sum_{c\in\{\text{TP}\}_{\alpha}} I\,oU(c) \quad (5)$$

where $IoU(c)$ is the intersection over union score between the predicted bounding box and ground truth bounding box which represent the true positive $c$. The detection accuracy at a threshold $\alpha$ is then calculated as:

$$\text{DetA}_{\alpha} = \frac{|\text{TP}_{\alpha}|}{|\text{TP}_{\alpha}|+|\text{FN}_{\alpha}|+|\text{FP}_{\alpha}|} \quad (6)$$

The association score for a single TP match $c$ between a detection and a ground truth (at threshold $\alpha$) is calculated as:

$$\mathcal{A}(c) = \frac{|\text{TPA}(c)|}{|\text{TPA}(c)|+|\text{FNA}(c)|+|\text{FPA}(c)|} \quad (7)$$

where corresponds to the number of true positive associations, corresponds to false negative associations and FPA corresponds to false positives associations. As shown in Figure, TPA, FNA and FPA are calculated (for each TP match $c$) over a predicted track and a ground truth track sequence. The association accuracy is then calculated as the average association score between all TPs matches in the dataset:

$$\text{AssA}_{\alpha} = \frac{1}{|\text{TP}_{\alpha}|}\sum_{c\in\{\text{TP}_{\alpha}\}} \mathcal{A}\,(c) \quad (8)$$

Lastly, HOTA, for a given threshold $\alpha$, can be defined as a combination of the above presented accuracy metrics:

$$\mathrm{HOTA}_{\alpha} = \sqrt{\mathrm{DetA}_{\alpha} \cdot \mathrm{AssA}_{\alpha}} = \sqrt{\frac{\sum_{c \in \{\mathrm{TP}_{\alpha}\}} \mathcal{A}(c)}{|\mathrm{TP}_{\alpha}| + |\mathrm{FN}_{\alpha}| + |\mathrm{FP}_{\alpha}|}} \tag{9}$$

Since DetA and AssA depend on the LocA values for a given threshold $\alpha$, HOTA is defined as the average $\mathrm{HOTA}_{\alpha}$ over a range of $\alpha$ thresholds:

$$\mathrm{HOTA} = \frac{1}{19} \sum_{\alpha \in \{0.05,\ 0.1,\ \ldots,\ 0.9,\ 0.95\}} \mathrm{HOTA}_{\alpha} \tag{10}$$

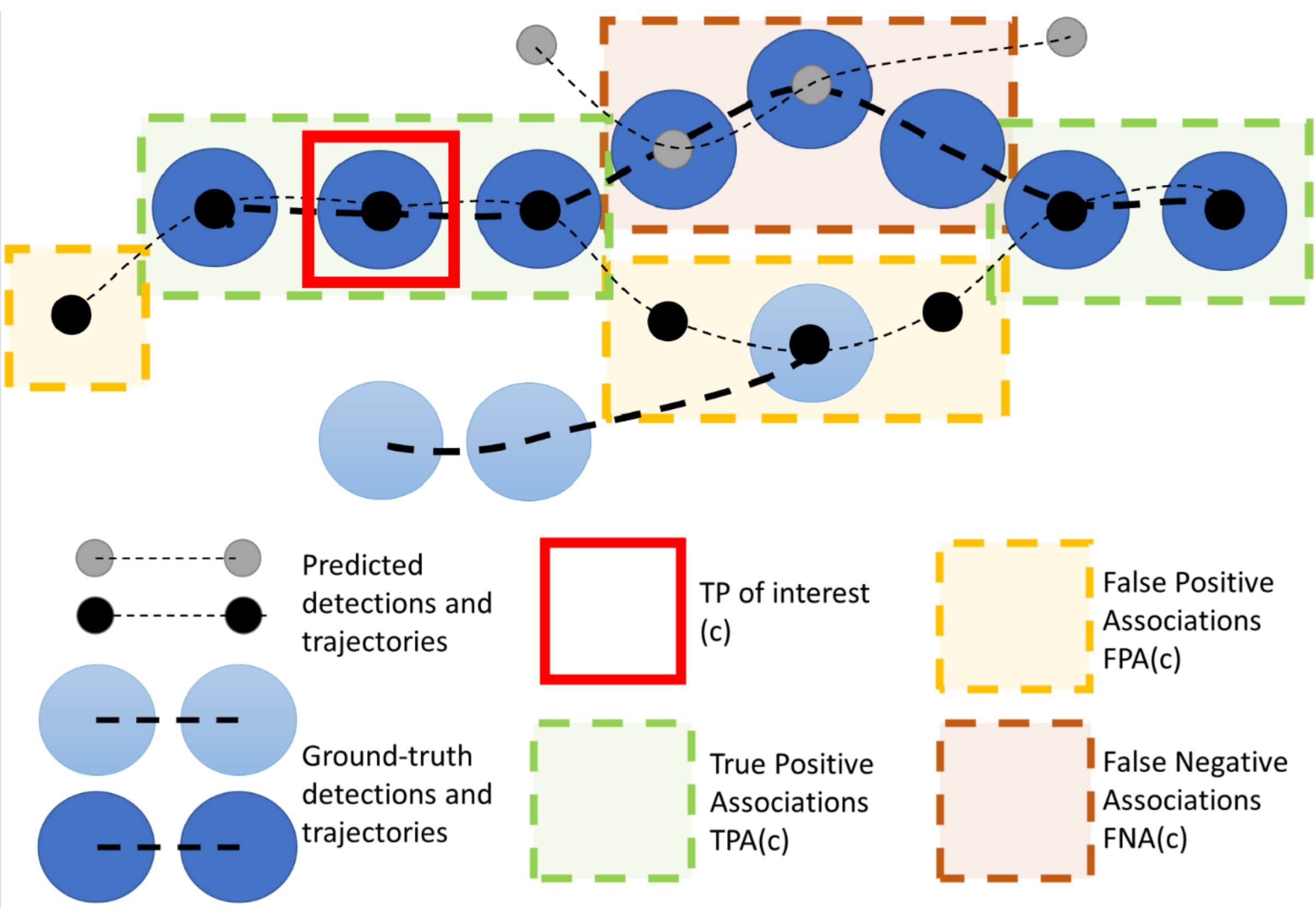

**Figure 6.** Example of predicted and ground truth tracks used to calculate $\mathcal{A}(c)$ over a single TP match $c$ (Luiten et al. 2021).

HOTA assumes an object whose trajectory can be defined by its position in a 2D image over time. Even though the data association is done fully in 3D, image space had to be used to evaluate HOTA. Therefore, the trajectory of a tomato track is defined as the positions in the image plane of the detected bounding boxes assigned to that tomato over a sequence of frames. This means that the relative position of the tomato with respect to the camera over the sequence is converted to a set of 2D image coordinates which can be thought as the projections of the 3D points of the tomato into the image plane over the sequence.

As shown in Figure 3, the environment is more challenging in the upper areas of the plant compared to the lower ones. Since the representation dataset was recorded in a semi-cylinder path with ten different height steps, the performance for all the metrics over these ten steps was evaluated. When the metric did not depend on previous viewpoints of the sequence, the performance of each step was calculated over the ten viewpoints of that specific height step. This is the case for the detection performance and the camera error analysis. When the performance referred to the output of the world model, where previous frames from both the same or different height steps were needed, the performance was evaluated using data from the first frame until the last frame of the specific height step. For every height step the tracking results of the seven plants were averaged.

# 3. Results

## 3.1 Pre-processing

As shown in Figure 5, the first step of the world modelling algorithm is to detect the object of interest in an image. As shown in Table 1, the trained Mask R-CNN model is able to achieve a recall of 0.86 and a precision of 0.71 when a confidence threshold of 0.5 is used. When a larger threshold is used, the recall decreases to 0.66 while the precision increases to 0.88. A larger recall was expected when a lower threshold was used, as more detections would be made. However, the chance of false positives also increases, reducing the precision. Figure 7 shows the precision and recall results per height step in the semi-cylinder path. On the one hand, it can be seen how the precision does not change significantly over different heights, although it slightly drops from the first half of the height steps to the second half. This means that there are few more false positives in the higher areas of the plant. On the other hand, recall drops steadily from the first until the last height, meaning that it is harder for the detection network to detect tomatoes in the upper parts of the plant.

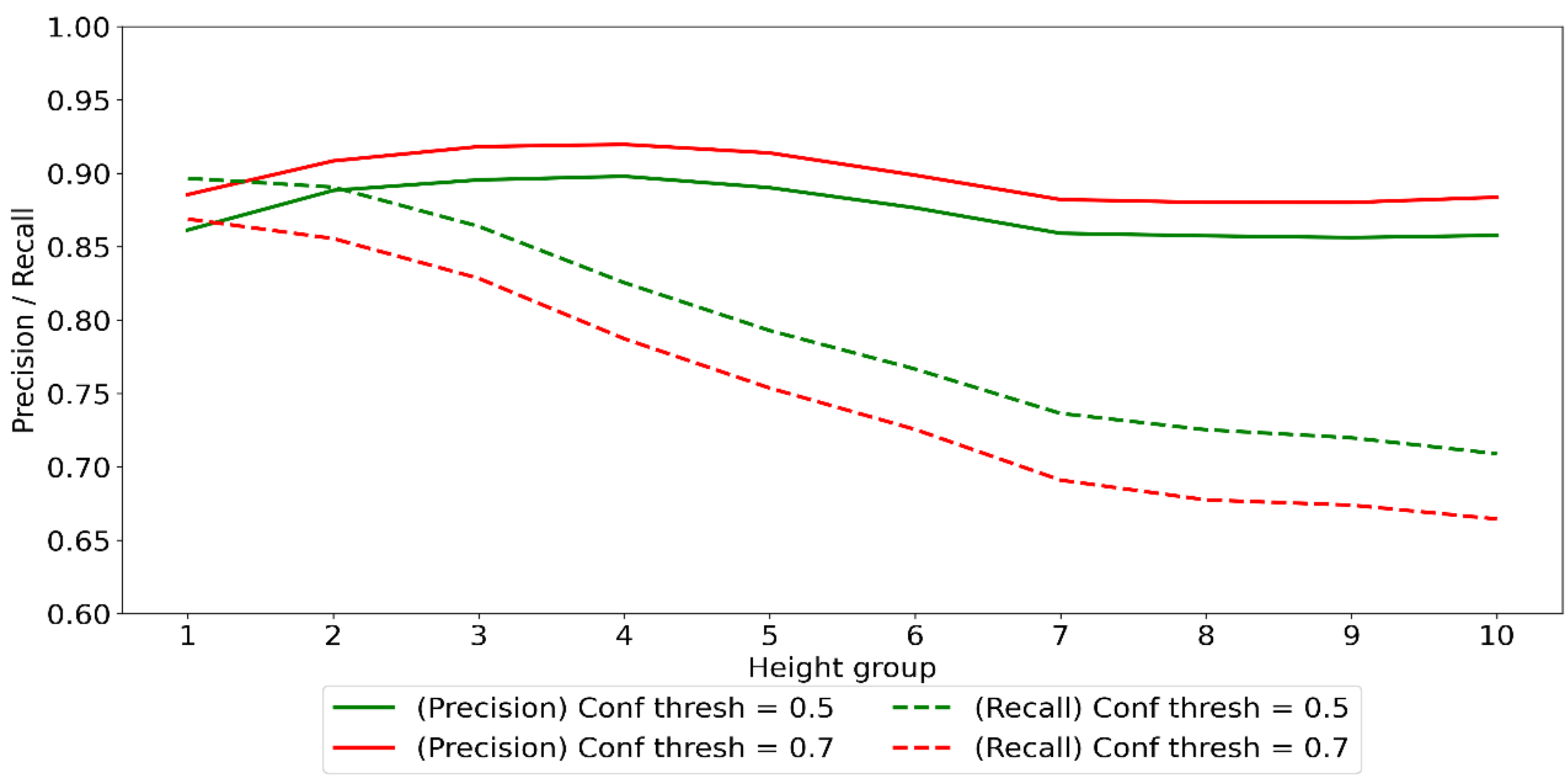

**Figure 7.** Precision and recall values for different confidence score thresholds using an IoU threshold of 0.5. The results are shown per height step in the semi-cylinder path and averaged over the seven recorded plants at every step.

**Table 1.** Precision and recall values for different confidence score thresholds using an IoU threshold of 0.5.

| Confidence threshold | Precision | Recall |
|---|---|---|
| 0.5 | 0.86 | 0.71 |
| 0.7 | 0.88 | 0.66 |

Table 2 shows the percentage of detections that are rejected due to a lack of 3D points in the structured point cloud (RD), and the percentage of detections whose sphere fitting results in non-valid centre or radius (NSF). It can be seen how both RD and NSF increase with the camera height independently of the detection confidence threshold. RD approximately doubles from the lowest steps to the highest steps, while NSF increases approximately between 3 and 5 percentage points.

In both Figure 7 and Table 2, it can be seen how the performance of both camera and detection worsens in the higher plants of the plant. This was expected, as the higher parts of a tomato plant contain more leaves creating more clutter and occlusions, and smaller and less ripe tomatoes, as illustrated in Figure 3.

**Table 2.** Percentage of rejected detections (RD) to which no points from the point cloud could be assigned, and percentage of detections with non-valid sphere fitting (NSF).

| Confidence threshold | 0.5 | | 0.7 | |
|---|---|---|---|---|
| Height step | RD | NSF | RD | NSF |
| 1 | 2.37 | 50.32 | 1.22 | 48.02 |
| 2 | 3.37 | 46.52 | 1.70 | 44.83 |
| 3 | 3.00 | 46.82 | 1.46 | 44.54 |
| 4 | 3.53 | 47.74 | 1.90 | 45.33 |
| 5 | 4.55 | 49.16 | 2.92 | 46.72 |
| 6 | 4.71 | 50.71 | 3.07 | 47.99 |
| 7 | 4.83 | 52.39 | 3.23 | 49.46 |
| 8 | 4.64 | 53.77 | 3.14 | 50.78 |
| 9 | 4.61 | 54.44 | 3.15 | 51.34 |
| 10 | 4.58 | 55.08 | 3.16 | 51.95 |

### 3.2. World model representation

Figure 8 shows the resulting world model representation over a plant point cloud. Figure 8a shows the resulting point cloud after merging several viewpoints taken from the data, and Figure 8b shows the 3D world model representation, where tomatoes are represented as individual spheres, over the plant point cloud. The representation is not perfect, as there are two false positives in the bottom left, and a missing tomato in the middle-top part of the figure. The proposed method can update the world model with a rate up to 10 Hz. However, more of the 95% of the time on average is spent in the pre-processing and detection step. The proposed tracking algorithm itself is simple and efficient. To further increase the update rate of the world model, faster detection algorithms, such as YOLACT++ (Bolya et al. 2022), can be used.

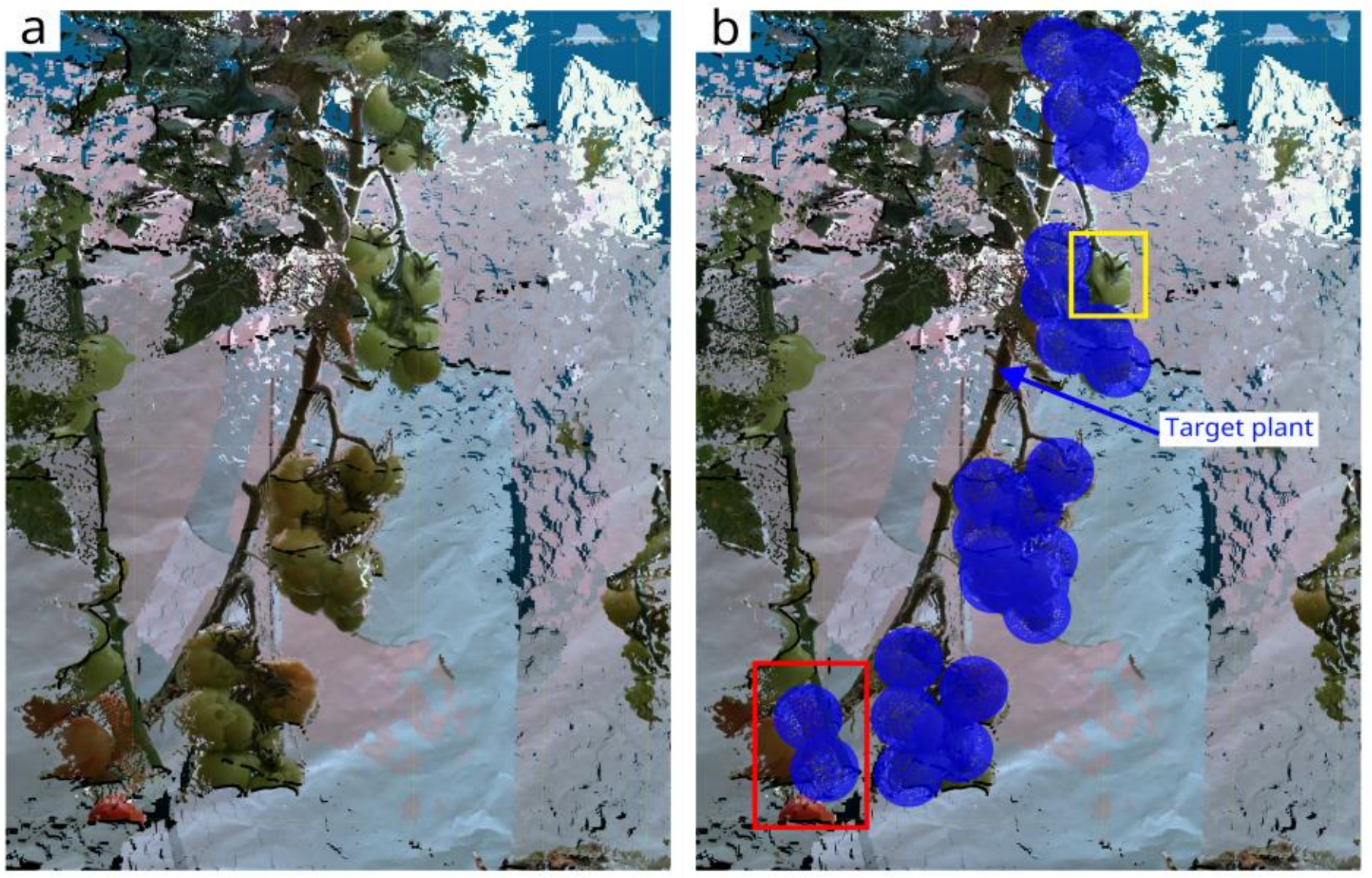

**Figure 8**. Example of the representation of a plant after the whole sequence is used. The left image (a) shows a rendered viewpoint of the combined point cloud of several sequential point clouds. The right image (b) shows the same viewpoint of the world model of the plant on top of the original point cloud, with the tomatoes represented by blue spheres. False positives, which are actual tomatoes, but belonging to another plant, are marked with a red box. False negatives are marked with a yellow box.

**Table 3.** Results of the total counting and tracking experiments. The performance is shown per height step in a cumulative manner. This means that each height step contains the data from all the previous steps. The height step marked with * corresponded with the whole sequence of frames. Numbers in bold correspond to the best values per metric.

| Confidence threshold | | 0.5 | | | | | 0.7 | | | | |
|---|---|---|---|---|---|---|---|---|---|---|---|
| Height step | $n_{init}$ | MAPE | MPE | HOTA | DetA | AssA | MAPE | MPE | HOTA | DetA | AssA |
| 1 | 0 | 17.48 | -13.86 | **71.47** | **68.82** | **74.78** | 13.50 | -8.28 | 69.49 | 68.44 | 71.03 |
| | 1 | 14.60 | -5.65 | 64.74 | 62.26 | 67.84 | 8.95 | 4.94 | 63.05 | 61.04 | 65.59 |
| 2 | 0 | 11.35 | -10.87 | 66.68 | 68.44 | 65.24 | 7.98 | -6.18 | 66.16 | 68.28 | 64.36 |
| | 1 | 9.82 | -3.44 | 63.34 | 64.62 | 62.38 | **5.06** | 3.52 | 62.44 | 63.68 | 61.50 |
| 3 | 0 | 17.79 | -17.53 | 63.50 | 67.47 | 59.98 | 13.12 | -9.87 | 64.45 | 66.99 | 62.24 |
| | 1 | 11.86 | -3.85 | 61.17 | 64.16 | 58.54 | 11.63 | 3.53 | 61.63 | 63.08 | 60.41 |
| 4 | 0 | 15.41 | -11.44 | 60.57 | 64.34 | 57.24 | 10.62 | -3.41 | 60.57 | 63.30 | 58.17 |
| | 1 | 14.21 | 1.28 | 58.49 | 61.55 | 55.80 | 12.70 | 8.61 | 58.31 | 59.77 | 57.08 |
| 5 | 0 | 18.50 | -11.56 | 58.72 | 61.86 | 55.96 | 11.60 | -0.70 | 58.52 | 60.68 | 56.66 |
| | 1 | 14.73 | 1.76 | 56.58 | 59.53 | 54.00 | 11.36 | 8.35 | 56.09 | 57.73 | 54.69 |
| 6 | 0 | 22.04 | -18.58 | 57.26 | 59.98 | 54.91 | 11.91 | -7.67 | 56.94 | 58.62 | 55.53 |
| | 1 | 15.39 | -0.37 | 55.45 | 57.85 | 53.39 | 10.92 | 7.02 | 54.58 | 55.80 | 53.57 |
| 7 | 0 | 23.94 | -20.61 | 55.44 | 57.79 | 53.42 | 12.78 | -11.30 | 55.07 | 56.24 | 54.15 |
| | 1 | 14.24 | -2.16 | 53.43 | 55.48 | 51.69 | 10.36 | 5.59 | 52.56 | 53.32 | 52.01 |
| 8 | 0 | 22.52 | -20.10 | 54.61 | 57.06 | 52.50 | 13.27 | -11.78 | 54.00 | 55.25 | 52.99 |
| | 1 | 13.17 | **-0.89** | 52.55 | 54.57 | 50.84 | 9.67 | 5.72 | 51.55 | 52.29 | 51.01 |
| 9 | 0 | 25.26 | -22.51 | 54.71 | 56.78 | 52.95 | 15.73 | -13.20 | 54.03 | 55.04 | 53.27 |
| | 1 | 14.51 | -3.59 | 52.63 | 54.28 | 51.27 | 9.56 | 3.96 | 51.62 | 52.12 | 51.31 |
| 10* | 0 | 27.50 | -23.86 | 54.32 | 55.98 | 52.96 | 18.01 | -13.88 | 53.53 | 54.39 | 52.90 |
| | 1 | 14.72 | -3.20 | 52.51 | 53.57 | 51.70 | 12.21 | 4.50 | 51.25 | 51.53 | 51.16 |

As table 3 shows, the columns of MAPE and MPE present on average better values when a confidence threshold of 0.7 is used. This suggests that a larger precision is more important than larger recall values. Due to the multi-view nature of the proposed algorithm, if a fruit is not detected in one viewpoint (frame), it can still be detected in one of the following viewpoints. Therefore, low recall values can be overcome with multi-view perception. On the contrary, a larger false positive rate, that corresponds with lower precision values, has the potential to inject non-existing fruits into the tracking system that are not removed further on in the current algorithm. Regarding $n_{init}$, the rows that indicate a threshold $n_{init} = 1$ have MAPE and MPE values closer to zero than their counter parts rows with $n_{init} = 0$. Similarly, this can be explained due to the multi-view nature of the algorithm. True positive detections that are not seen consecutively over 2 frames and, therefore, do not pass the threshold $n_{init} = 1$, can still appear consecutively in later frames and be added to the world model; while false positives are filtered. When $n_{init} = 0$ both TPs and FPs that are seen only once are added to the world model. Consequently, the configuration that allows for more false positives, a confidence threshold of 0.5 and $n_{init}$ of 1, have higher MAPE values, and presents the largest overestimation of tomatoes.

HOTA, DetA and AssA show lower performance in the higher parts of the plant, probably due to the more challenging conditions. The same happens with MAPE when $n_{init}$ is set to 0, as false positives are more likely to appear in the top parts of the plant. Even though recall values decline noticeably in the higher parts of the plant, when $n_{init}$ is set to 1, MAPE values do not increase as a result. This again could be prevented by multi-view perception, as although the chance of detecting a tomato is lower, there are still multiple frames where the tomato can be detected. MPE performance does not decrease either over height because under- and over-counting results between the different plants can cancel each other out, producing values close to zero.

While comparing MAPE and HOTA performance over different parameters and height steps, it can be seen how similar MAPE values can have very different HOTA results. For instance, the MAPE results of 11.35 at step 2, $n_{init} = 0$, and a confidence threshold of 0,5; and 11.36 at step 5, $n_{init} = 1$ and a confidence threshold of 0.7, have HOTA values of 66.68 and 56.09 respectively. This is due to how HOTA is calculated, since, with more frames of the recorded sequences, the number of predicted and ground truth tracks increases, which consequently increases the chances of association and detection errors. This is especially true in the used data, as the later frames are more complex than the frames from the bottom part of the plant. Furthermore, the best MAPE results are found with parameters that reduce the number of false positives in spite of a lower detection rate, confidence threshold of 0.7 and $n_{init}$ of 1. However, HOTA and its subsequent metrics, yield better results when the opposite is true, i.e. detection threshold of 0.5 and $n_{init} = 0$, which increases the detection rate and the number of false positives. This can only be explained if a higher detection rate generated more TPs than FPs, circumstance that increases DetA, AssA and HOTA. Given the equations used to calculate DetA and $\mathcal{A}(c)$, a 1:1 increase ratio in TPs and FPs would result in a lower accuracy.

## 4. Discussion

The concept of a generic representation of the environments that can be used by robots across many tasks with a few modifications, like the objects of interest, has been proposed by several authors (Elfring et al. 2013; Persson et al. 2020; Wong et al. 2015; Inceoglu et al.

2019). However, their work was limited to controlled lab environments and did not evaluate in-depth the performance of their data association algorithms with metrics like the ones proposed by Luiten et al. (2021). In this work, it was showed how a generic representation, based on data association, can accurately represent and locate most objects in a real world tomato greenhouse environment. Nonetheless, the performance of the world model can be potentially improved in several steps of the pipeline.

The performance of the detection algorithm varied greatly between different height steps. In the lower part of the plants, the performance is similar to what other research has found in datasets that aim toward the de-leafed lower parts of tomato plants (Afonso et al. 2019). However, detection in the upper parts of the plant was more challenging, due to more leaves and smaller and less ripe tomatoes, leading to a lower performance. Using training data containing the tomato variety that is used in the counting and tracking experiment could potentially improve the detection performance. Furthermore, the quality of the point cloud affected the performance of both detection and tracking algorithms. Up to 4.58% of the detections were rejected due to a lack of 3D points, which affected precision and recall values. The predicted sphere of up to 55.08% of the detected tomatoes was non-valid, mainly due to a partial tomato point cloud with few points or with high noise levels. Although the impact of varying lighting conditions was not assessed in this study, it is an important factor to consider in future work as lighting conditions can greatly vary in agro-food environments. The Mask RCNN performance may be affected by these conditions. To mitigate this, Mask RCNN was trained on a dataset that included images captured under different lighting and environmental conditions, which probably improved the robustness to differences in illumination. Furthermore, the LiDAR sensor may be influenced by environmental light, leading to noisier point clouds and an increase in the number of rejected detections or invalid spheres.

The performance of the detection algorithm greatly affects the accuracy of the 3D tracking algorithm. This is something that Bewley et al. (2016) pointed out for Hungarian-based tracking algorithms. Concretely, the performed experiments showed that configurations which increase the recall and reduce the precision of the detection system decreased the counting performance while increasing the tracking performance. A cause for this is the inability of the world model to remove existing objects, which harms its ability to remove false positive detections. Consequently, false positives stay in the world model increasing the counting error. However, removing false positive tracks from the world model in an environment where occlusions are common is not an easy task. If an object is in the field of view of the camera, but it is not detected, it does not exclusively mean that it is a false positive. It could be occluded behind another object, like a leaf for example.

Localising and counting fruits in agro-food environments is a challenging task due to the complexity of the environment, yet it is key for robotic applications. The proposed algorithm can potentially yield a MAPE of 5.06%. Recently, Kirk et al. (2021) has shown the potential of similar tracking system. They were able to count strawberries with a MAPE of 3%;

although their conditions were more favourable with fewer occlusions and a larger distance between individual fruits. In similar environment conditions as the ones in this work, Halstead et al. (2021) developed and evaluated another similar tracking system to count bell peppers in individual plants, and achieved a MAPE of 4.1%. Their data also contained clutter and de-leafed areas of plants, but, since bell peppers do not grow in cluster like tomatoes, the distance between individual fruits is more favourable toward tracking. Smitt et al. (2021) extended Halstead et al. (2021) work to build an autonomous system that could count bell peppers along a row of plants in the greenhouse. Besides, they added bounding box re-projection between different frames using wheel odometry and the point cloud from the camera. They achieved a MAPE of 21% on tomato and sweet pepper plants. However, the tracking algorithm of all these works was 2D-based, and they used a fixed camera array which moved parallel to the plants' row. On the one hand, fixed camera arrays provide faster information than a camera attached to a robot arm. On the other hand, they are limited in the information they can offer as there are viewpoints that cannot be achieved. Consequently, they are less capable of overcoming certain occlusions. Furthermore, a 2D-based tracking system is at a clear disadvantage against a 3D-based one; as an object that appears in front of a previously seen object and occludes it will have very similar 2D camera coordinates to the occluded object.

In the conducted experiments, the counting error was evaluated to assess the performance of the representation by a 3D tracking algorithm. This is a common practice in the field (Smitt et al. 2021; Halstead et al. 2021; Kirk et al. 2021). However, from the experiments, it becomes evident that counting and tracking performance do not always yield similar results. A tracking algorithm can carry other types of errors that are not covered by counting performance metrics. For instance, during tracking, two or more objects can switch IDs during one or multiple frames. This might not affect the final count prediction, and is, therefore, not visible in the counting metrics. Furthermore, good counting results can originate from a set of detections which contain missed objects which are compensated by false positives. This type of errors harms the accuracy of the plant representation and might not be desired for some robotics and monitoring applications. Multi-object tracking (MOT) metrics such as the ones proposed by Luiten et al. (2021), give more insights into the actual tracking performance and should be used to properly evaluate the performance of any tracking algorithm.

The HOTA and AssA values at the higher viewpoints of the plant suggest that there are still numerous association errors. This is probably due to the simplicity of the tracking algorithm, where only the estimated Cartesian centre of the objects is used for tracking. Several multi-object tracking algorithms have shown how using other features in the data association step (Wojke et al. 2017) or an end-to-end approach (Meinhardt et al. 2022) perform better in scenarios with many occlusions.

All the data used was recorded following a semi-cylinder path around a plant, therefore, the viewpoints are not optimised for information gain and several images do not contain any valuable information at all, like Figure 3a. A NBV algorithm like the one proposed by (Wu et al. 2019; Burusa et al. 2022) has the potential to improve the efficiency and quality of the world model representation, as the viewpoints selected by the NBV algorithm might contain fewer occlusions of the relevant objects in the scene.

## 5. Conclusions and future work

In this study, a generic approach for 3D representation in robots, based on dynamic properties of individual objects, was introduced. This approach can generate precise representations in extremely occluded environments. Further development and assessment were carried out on a 3D MOT algorithm, which maintains and updates the objects in the representation. The world model was demonstrated to successfully locate and depict tomatoes in a challenging, densely cluttered real-world tomato greenhouse, with a high degree of occlusion, yielding a tracking accuracy of up to 71.47% using the HOTA metric and a low counting error of 5.06%. These findings have practical implications for applications such as fruit harvesting. The assessment using tracking metrics alongside counting metrics also revealed that the MOT algorithm's total count of objects at the end of a sequence inadequately describes its performance and fails to detect certain types of errors, particularly false positives and false negatives that offset one another. The primary source of errors in the MOT algorithm was identified by evaluating different stages of the pipeline and utilising appropriate and current metrics. These errors were found to originate from the detection and association performance in the most heavily cluttered regions of the plant.

To further improve the world model in future work, exploration of different deep learning methods to improve the tracking performance, such as Wojke et al. (2017) and Meinhardt et al. (2022), will be conducted to enhance the association performance. Some of these methods might be able to remove false positive objects out of the world model in the presence of occlusions due to being end-to-end deep learning approaches. To increase the detection performance, NBV algorithms can select viewpoints that contain fewer occlusions and provide the detection algorithm with images and point clouds easier to process. Therefore, the performance of the world model representation when it is coupled with a NBV algorithm will be evaluated.

## CRediT author statement

David Rapado-Rincon: Conceptualisation, Methodology, Software, Investigation, Data Processing, Writing - Original draft; Eldert J. van Henten: Conceptualisation, Writing - Review & Editing, Supervision, Funding acquisition; Gert Kootstra: Conceptualisation, Writing - Review & Editing, Supervision, Funding acquisition.

## Funding

This research is part of the project Cognitive Robotics for Flexible Agro-food Technology (FlexCRAFT), funded by the Netherlands Organisation for Scientific Research (NWO) grant P17-01.

## Declaration of competing interest

It is declared that there are no personal and/or financial relationships that have inappropriately affected or influenced the work presented in this paper.